\documentclass[11pt]{article}

\usepackage{acl}

\usepackage{times}
\usepackage{latexsym}
\usepackage[T1]{fontenc}
\usepackage[utf8]{inputenc}
\usepackage{microtype}
\usepackage{inconsolata}
\usepackage{graphicx}
\usepackage{booktabs}
\usepackage{amsmath,amssymb}

\title{Multilingual Sentence Embeddings for Linguistic-Integrated Reliability Audit}

\author{
  Ummugul Bezirhan, Ji Yoon Jung, Matthias von Davier \\
  Boston College \\
  Chestnut Hill, Massachusetts, USA \\
  \texttt{\{bezirhan, jungjg, vondavim\}@bc.edu}
}

\begin{document}
\maketitle

\begin{abstract}
Multilingual assessment systems commonly rely on translation for scoring and quality-control processes. We evaluate whether multilingual sentence embeddings can replace translated English input for Linguistic-Integrated Reliability Auditing (LiRA) across 11 PIRLS constructed-response items and three embedding models. Native-language embeddings reproduced translation-based reliability estimates closely while recovering responses excluded after translation failure, with no meaningful change in reliability.
\end{abstract}

\section{Introduction}
International large-scale assessments (ILSAs) such as PIRLS and TIMSS increasingly explore and implement automated approaches for scoring and quality control of constructed-response (CR) items (e.g., \citealp{jung2024combining, tyack2024cnn}). Because these assessments operate across hundreds of languages simultaneously, multilingual processing remains a central methodological challenge. Automated scoring systems must therefore support reliable semantic processing across linguistically diverse response sets while maintaining comparability across countries and languages. 
A common strategy for multilingual automated scoring is the translate-then-score paradigm, in which student responses are translated into English prior to scoring \citep{horbach2024crosslingual,jung2024combining}. This approach simplifies deployment by allowing a single English-centric scoring system to be applied across multiple languages and countries. More recent large language model (LLM)-based scoring frameworks have further integrated translation directly into prompting workflows, enabling multilingual responses to be processed within a unified scoring framework \citep{jung2025scalability}. Although these approaches have substantially improved scalability and multilingual coverage, they continue to rely on translation as an intermediate representation. Even when scoring is performed directly on responses written in their native language \citep{jung2026promptcompression}, translation often remains a critical component retained for downstream processes such as quality-control procedures, consistency checks, reliability analyses, and human review. While LLM-based context-aware translation better preserves the original meaning and nuances than traditional neural machine translation \citep{jung2025scalability}, any distortions introduced during translation may propagate beyond scoring into subsequent measurement and validation activities. 

Translation, however, is not necessarily a neutral preprocessing step. Prior research in cross-lingual representation learning has shown that semantic spaces are not fully isomorphic across languages, particularly for structurally distant or low-resource languages \citep{soegaard2018limitations,beinborn2020semantic}. Even when translations appear semantically accurate at the surface level, translation may alter local neighborhood structure, pairwise similarity relationships, or embedding geometry. Such distortions may be especially important for downstream tasks that rely directly on semantic similarity relationships among responses \citep{bezirhan2026lost}. 

This issue is particularly relevant for LiRA (Linguistic-integrated Reliability Audit; \citealp{jung2025reconceptualizing}), a recently proposed reliability auditing framework for multilingual scoring systems. LiRA estimates reliability using semantic similarity relationships among student responses. For each response, the framework retrieves the most semantically similar neighboring responses in the embedding space and constructs a similarity-weighted benchmark score using the neighboring responses' labels. Reliability is then estimated by assessing the agreement between the original scores and the neighborhood-derived benchmark scores. Because LiRA operates entirely in embedding space, translation-induced changes in semantic neighborhoods may affect neighbor retrieval, benchmark construction, and ultimately, reliability estimates. 

Recent work has demonstrated that translation can introduce measurable semantic drift in multilingual embedding spaces. Translation-induced shifts have been observed in pairwise similarity, local neighborhood preservation, and global geometric alignment across multilingual sentence embedding models \citep{artetxe2019massively,conneau2019xlm,kornblith2019similarity}. However, it remains unclear whether these geometric changes meaningfully affect downstream reliability auditing tasks such as LiRA. Importantly, advances in multilingual embedding models now allow semantic representations to be computed directly from native-language responses without requiring translation into English. 

The present study investigates whether translation is necessary for LiRA. Specifically, we compare an original translation-based LiRA pipeline against a multilingual pipeline that embeds native-language responses directly using multilingual sentence embeddings. Using 11 PIRLS CR items across multiple languages, we evaluate whether multilingual LiRA can reproduce reliability estimates from translated English responses while reducing reliance on translation preprocessing and recovering responses excluded due to missing translations. 

\section{Conceptual Framework} 
Translation is not merely a technical preprocessing step; it can alter the linguistic and semantic characteristics of student responses. Machine translation and LLM-based translation systems are designed to produce fluent target-language output, which may normalize features commonly observed in student writing, including fragmentary expressions, nonstandard spelling, and code-switching. In some cases, translation may also alter or omit information contained in the original response, particularly when responses are short, ambiguous, or context-dependent \citep{beinborn2020semantic,somers2006detecting}. Such changes are especially relevant for short CR items, where a single idea or phrase may determine the assigned score. Even minor shifts in meaning may therefore influence both automated scoring outcomes and downstream reliability analyses. 

These challenges are not distributed uniformly across languages. Translation quality is generally lower for low-resource and morphologically complex languages, where limited training data, rich inflectional systems, and greater orthographic variation increase the likelihood of translation errors \citep{beinborn2020semantic}. In operational settings, translation may also be bypassed or yield incomplete output, necessitating the exclusion of affected responses from subsequent analyses. As a result, translation-based pipelines may disproportionately affect the very languages and response types that are already most difficult to model reliably. 

Translation-induced semantic drift does not necessarily arise from outright mistranslation. Even when a translation is linguistically accurate, the translated expression may not carry exactly the same semantic associations as the original. Research in cross-lingual semantics has shown that concepts often differ in semantic scope across languages \citep{conneau2018xnli}, and that projecting languages into a shared representation space can either sharpen fine-grained distinctions through cross-lingual sense disambiguation \citep{faruqui2014improving} or blur them through phenomena such as hubness, where a small number of representations appear disproportionately often among the nearest neighbors of others \citep{dinu2014hubness}. Consequently, responses that appear equivalent to human readers may nevertheless occupy different locations within a semantic space and exhibit different neighborhood relationships. For retrieval-based methods that rely on semantic similarity, these local changes may be more consequential than shifts in global geometric structure. For embedding-based methods, these differences matter because they affect not only the representation of individual responses but also their relationships to surrounding responses \citep{bezirhan2026lost}. A translated response may remain broadly similar to its original version while exhibiting different nearest-neighbor relationships, altered local density, or increased hubness. Such changes are particularly relevant for retrieval-based methods, which depend on local neighborhood structure rather than global geometric alignment. 

LiRA estimates reliability by leveraging semantic relationships among responses in an embedding space rather than relying on additional human scoring \citep{jung2025reconceptualizing}. For each response, LiRA retrieves the most semantically similar neighboring responses, constructs a similarity-weighted benchmark score from their labels, and estimates reliability through weighted exact agreement between the original score and this neighborhood-derived benchmark. Because LiRA operates entirely on semantic neighborhoods, its validity depends directly on the quality of the underlying representation space. In translation-based implementations, this space is constructed from English translations of multilingual responses. Consequently, translation-induced changes in local neighborhood structure may influence neighbor retrieval, benchmark construction, and ultimately reliability estimates. 

Recent multilingual sentence embedding models provide an alternative by representing responses directly in their native language within a shared multilingual semantic space. If multilingual representations preserve the local neighborhood structure required for benchmark construction, translation may be unnecessary for semantic-similarity-based reliability auditing. The present study evaluates this possibility by comparing translation-based and multilingual LiRA pipelines under otherwise identical conditions.

\section{Methods}
\subsection{Data and Design}

We use PIRLS 2021 reading-comprehension responses to 11 dichotomously scored CR items drawn from three reading passages. The number of validly scored responses ranges from 12,729 to 15,794 per item, spanning the 29 languages of the participating PIRLS 2021 countries.

The objective of the study was to isolate the effect of translation on LiRA reliability estimates. To achieve this, the LiRA procedure and embedding model were held constant, with only the textual representation provided to the encoder varied. Two processing conditions were evaluated. In the translation-based condition, the responses were translated into English using GPT-4.1 with a context-rich translation framework prior to embedding. This corresponds to the original LiRA implementation \citep{jung2025reconceptualizing}. For the proposed translation-free condition, the responses were embedded directly in their original language without a translation step.

In both conditions, the same embedding model, LiRA parameters, and reliability estimation procedure were used. Consequently, any differences in reliability estimates can be attributed to the use of translated versus native-language representations rather than differences in model architecture or scoring methodology.

\subsection{The LiRA Procedure}

We summarize LiRA formally to make explicit where translation can enter the reliability-estimation process. Consider a set of $N$ scored responses to a single item. Each response $i$ has an observed human score $h_i \in S$, where $S$ is the item's score set. In the present study, all items are dichotomously scored, so $S=\{0,1\}$.

Let $x_i$ denote the text representation of response $i$. In the translation-based condition, $x_i$ is the English translation of the response; in the translation-free condition, $x_i$ is the original native-language response. An encoder $f$ maps the response text to an embedding,
\begin{equation}
\mathbf{e}_i = f(x_i) \in \mathbb{R}^{d},
\label{eq:embedding}
\end{equation}
and similarity between two responses is computed using cosine similarity,
\begin{equation}
\operatorname{sim}(i,j)
=
\frac{\mathbf{e}_i \cdot \mathbf{e}_j}
{\lVert \mathbf{e}_i \rVert \, \lVert \mathbf{e}_j \rVert}.
\label{eq:cosine-similarity}
\end{equation}

For each response $i$, LiRA retrieves the set $N_k(i)$ containing the $k$ most similar responses to $i$, excluding the response itself; here $k=3$, following the original LiRA implementation. Within this neighborhood, the total similarity weight supporting each candidate score $s \in S$ is
\begin{equation}
w_i(s)
=
\sum_{\substack{j \in N_k(i) \\ h_j=s}}
\operatorname{sim}(i,j),
\label{eq:score-weight}
\end{equation}
and the similarity-weighted majority score is
\begin{equation}
s_i^{*}
=
\arg\max_{s \in S} w_i(s).
\label{eq:majority-score}
\end{equation}

LiRA assigns this benchmark score only when the winning score receives a sufficient share of the total neighborhood weight. Let
\begin{equation}
p_i
=
\frac{w_i(s_i^{*})}
{\sum_{j \in N_k(i)} \operatorname{sim}(i,j)}.
\label{eq:winning-share}
\end{equation}

If $p_i$ exceeds the threshold $\tau$, the benchmark score $s_i^{*}$ is assigned; otherwise, response $i$ is flagged as inconsistent. Following the original implementation, we use $\tau=0.60$. Responses flagged as missing or untranslated are excluded from the translation-based reliability calculation, and responses flagged as meaningless are assigned the lowest score category. For each response assigned a numeric benchmark score, define the average cosine similarity of its retrieved neighbors,
\begin{equation}
\bar{c}_i
=
\frac{1}{k}
\sum_{j \in N_k(i)}
\operatorname{sim}(i,j).
\label{eq:average-similarity}
\end{equation}

Let $V$ denote the set of valid responses with both an observed human score and an assigned numeric LiRA benchmark. Weighted Exact Agreement (WEA) is then
\begin{equation}
\operatorname{WEA}
=
\frac{
\sum_{i \in V}
\bar{c}_i \,
\mathbb{I}(h_i=s_i^{*})
}{
\sum_{i \in V}
\bar{c}_i
}.
\label{eq:wea}
\end{equation}

Responses with higher average neighborhood similarity therefore contribute more heavily to the reliability estimate. This formalization makes clear where translation can affect LiRA. All downstream quantities are functions of the response embedding.

\subsection{Multilingual Embedding Models}

To assess the robustness of the proposed approach across embedding architectures, we evaluated three multilingual sentence encoders representing different modeling objectives: LaBSE \citep{feng2022labse}, which was designed for cross-lingual alignment; Multilingual-E5 \citep{wang2024multilingual}, which was optimized for multilingual retrieval tasks; and Qwen3-Embedding \citep{zhang2025qwen3embedding}, a general-purpose multilingual embedding model. For each encoder, we constructed both translation-based and translation-free LiRA pipelines using identical retrieval, voting, and reliability-estimation parameters. This design isolates the effect of replacing translated responses with native-language responses while holding the underlying encoder and all other components of the LiRA framework constant.

\subsection{Evaluation}

Because GPT-4.1 translation occasionally fails or produces untranslated output, the translation-based and translation-free pipelines do not always operate on identical response sets. To separate translation effects from sample-composition effects, we conducted both matched-sample and full-sample analyses. In addition, we recorded the number of rescued responses, defined as responses excluded because of translation failure but retained in the translation-free pipeline. To assess the stability of LiRA outputs across conditions, we also computed prediction agreement, defined as the proportion of responses that receive the same benchmark score under both pipelines, and inconsistency-flag agreement, defined as the proportion of responses that receive the same inconsistency classification.

\section{Results}

Table~\ref{tab:encoder-results} summarizes LiRA reliability estimates obtained from the translation-based and translation-free pipelines across the three multilingual embedding models. Across all encoders, reliability estimates were highly similar between conditions. Mean WEA differences ranged from $-0.0030$ for LaBSE to $-0.0086$ for Multilingual-E5, with all encoder-level differences remaining less than $0.01$. These results indicate that translation-free LiRA reproduces translation-based reliability estimates consistently across multilingual embedding architectures. The findings therefore suggest that the observed equivalence is not specific to a particular encoder design but reflects a broader robustness of the LiRA framework to the use of native-language representations.

\begin{table}[t]
\centering
\resizebox{\columnwidth}{!}{%
\begin{tabular}{lcccc}
\toprule
\textbf{Encoder} &
\textbf{WEA} &
\textbf{WEA} &
\textbf{$\Delta$WEA} &
\textbf{Prediction} \\
&
\textbf{English} &
\textbf{Multilingual} &
&
\textbf{Agreement} \\
\midrule
Qwen3            & 0.9531 & 0.9487 & $-0.0045$ & 0.9525 \\
Multilingual-E5  & 0.9486 & 0.9400 & $-0.0086$ & 0.9467 \\
LaBSE            & 0.9501 & 0.9471 & $-0.0030$ & 0.9572 \\
\bottomrule
\end{tabular}%
}
\caption{Reliability and cross-pipeline agreement by encoder, averaged across 11 items.}
\label{tab:encoder-results}
\end{table}

Item-level differences were small throughout, as shown in Table~\ref{tab:item-delta}. For every encoder, all 11 items fell within about one and a half percentage point of the translated baseline, and the largest single-item change across all encoders was $-0.0142$ (E5). The slight ordering that does appear, LaBSE tracking the baseline most closely and E5 least closely, with Qwen between, is consistent with the encoders' design objectives, but the spread is small enough that encoder choice is not consequential for the conclusion. Because the encoders produced closely similar results, we use Qwen3-Embedding, a strong general-purpose multilingual encoder that also attained the highest absolute WEA in both conditions, for the item-level analyses that follow.

\begin{table*}
\centering
\begin{tabular}{lrrrr}
\toprule
\textbf{Item} &
\textbf{Ref. WEA} &
\textbf{$\Delta$ LaBSE} &
\textbf{$\Delta$ Qwen} &
\textbf{$\Delta$ E5} \\
\midrule
Item 1  & 0.9869 & $-0.0007$ & $-0.0014$ & $-0.0059$ \\
Item 2  & 0.9791 & $-0.0013$ & $-0.0047$ & $-0.0063$ \\
Item 3  & 0.9885 & $-0.0046$ & $-0.0088$ & $-0.0122$ \\
Item 4  & 0.9940 & $-0.0029$ & $-0.0030$ & $-0.0076$ \\
Item 5  & 0.9945 & $-0.0017$ & $-0.0033$ & $-0.0052$ \\
Item 6  & 0.9247 & $-0.0039$ & $+0.0021$ & $-0.0058$ \\
Item 7  & 0.9481 & $-0.0085$ & $-0.0040$ & $-0.0140$ \\
Item 8  & 0.9169 & $-0.0036$ & $-0.0111$ & $-0.0125$ \\
Item 9  & 0.9630 & $-0.0021$ & $-0.0044$ & $-0.0048$ \\
Item 10 & 0.8957 & $-0.0037$ & $-0.0068$ & $-0.0142$ \\
Item 11 & 0.8932 & $+0.0002$ & $-0.0039$ & $-0.0060$ \\
\midrule
Mean    & 0.9531 & $-0.0030$ & $-0.0045$ & $-0.0086$ \\
\bottomrule
\end{tabular}
\caption{Per-item $\Delta$WEA by encoder. Ref.\ WEA is the Qwen translation-based baseline shown for orientation. Each encoder's $\Delta$WEA is calculated against its own translation-based baseline.}
\label{tab:item-delta}
\end{table*}

\begin{figure*}[t]
    \centering
    \includegraphics[width=\textwidth]{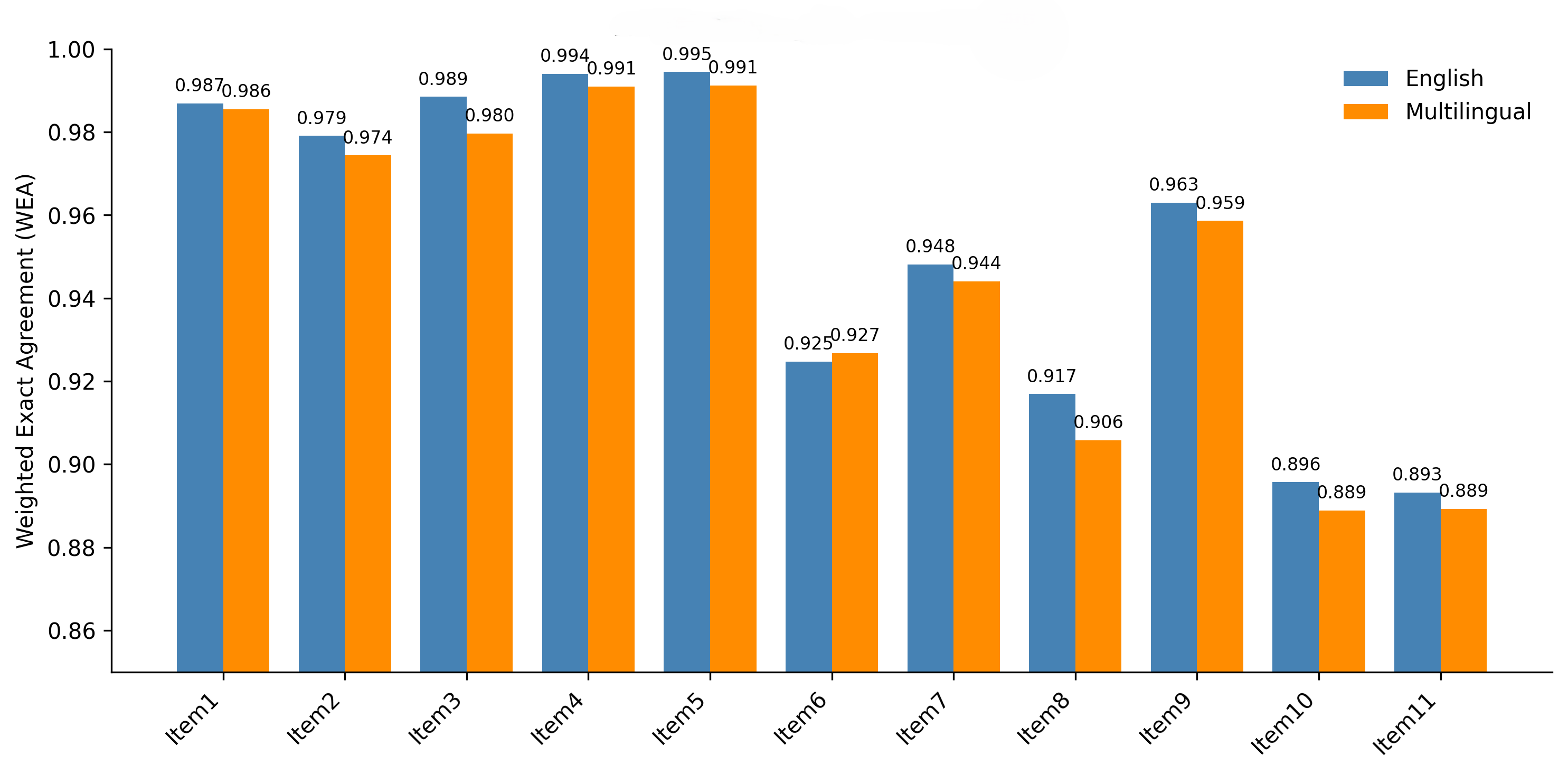}
    \caption{LiRA WEA by item, English vs.\ multilingual Qwen embeddings.}
    \label{fig:qwen-wea}
\end{figure*}

Figure~\ref{fig:qwen-wea} shows the per-item translated and native WEA for the Qwen pipeline. Across all 11 items, the two conditions are close, with native WEA within about one percentage point of translated WEA. The largest item-level gaps occur on the lower-agreement items (Item 8, 0.917 vs.\ 0.906; Item 10, 0.896 vs.\ 0.889), while on one item (Item 6) the native pipeline slightly exceeds the translated one (0.925 vs.\ 0.927). The pattern is one of close correspondence rather than uniform degradation.

Although reliability estimates were highly similar across conditions, the translation-based and translation-free pipelines did not always produce identical benchmark assignments. For Qwen3, benchmark agreement averaged 95.3\% across items, indicating that approximately 95\% of responses received the same LiRA benchmark score under both pipelines. Agreement in inconsistency classifications was effectively perfect across items, exceeding 99.9\% in all cases.

The combination of high benchmark agreement and small $\Delta$WEA values suggests that LiRA reliability estimates are robust to moderate changes in local semantic neighborhoods. While translation and native-language embeddings may occasionally retrieve different neighboring responses, these differences rarely alter the overall reliability conclusions produced by the framework.

Across the 11 items, 384 responses (13--66 per item) failed GPT-4.1 translation and were dropped from the translation-based pipeline. The multilingual pipeline recovered all of them under Qwen and E5, and 383 of 384 under LaBSE (a single response LaBSE could not embed). Including the rescued responses left reliability essentially unchanged; WEA on the full set differed from the matched set by at most 0.0004 per item for every encoder, so the coverage gain comes without any measurable reliability cost. Removing translation, therefore, makes the audit more inclusive: it scores responses that translation would discard, which are usually concentrated in low-resource and morphologically complex languages.

\section{Discussion}

The purpose of this study was to evaluate whether translation is necessary for LiRA-based reliability auditing in multilingual assessment systems. Across 11 PIRLS CR items, three multilingual embedding models, and more than 160,000 scored responses, translation-free LiRA produced reliability estimates that were highly similar to those obtained using translated English responses. Mean differences in WEA remained below 1 percentage point for all embedding models, benchmark agreement exceeded 94\% in every condition, and responses excluded due to translation failure were successfully recovered without altering reliability estimates.

These findings indicate that multilingual sentence embeddings can reproduce translation-based LiRA reliability estimates with only minimal differences, and this conclusion is robust across encoder architectures. Although LaBSE, Multilingual-E5, and Qwen3-Embedding were developed with different objectives and training strategies, all three produced highly similar patterns of results. The consistency of the findings suggests that the observed equivalence reflects a property of the LiRA framework itself rather than the behavior of any particular embedding model.

The results also provide insight into the relationship between semantic drift and reliability auditing. Prior work has shown that translation can alter semantic relationships, local neighborhood structure, and global embedding geometry (e.g., \citealp{conneau2018xnli,bezirhan2026lost}). From a theoretical perspective, such changes could affect LiRA because benchmark scores are constructed directly from nearest-neighbor relationships. However, the present findings indicate that moderate changes in semantic neighborhoods do not necessarily translate into meaningful differences in reliability estimates. Although approximately five percent of benchmark assignments differed between the translation-based and translation-free pipelines, the resulting changes in WEA were negligible. This suggests that LiRA is relatively robust to modest perturbations in local semantic structure.

The findings also have practical implications for multilingual assessment systems. Translation has traditionally served as a mechanism for standardizing multilingual responses into a common language before scoring and downstream analysis. While this approach simplifies processing, it introduces additional computational cost, dependence on translation quality, and the possibility of excluding responses that cannot be translated successfully. The present results demonstrate that multilingual embeddings provide a viable alternative for reliability auditing.

Several limitations should be acknowledged. First, the study focused on dichotomously scored PIRLS reading-comprehension items. Whether similar results hold for longer responses, polytomous scoring schemes, or other assessment domains remains an open question for future research. Second, although three multilingual embedding models were evaluated, the study did not examine the full range of contemporary multilingual encoders. Future research should investigate whether the observed robustness extends to additional model families and emerging multilingual representation methods. Finally, the present analysis focused on reliability estimation rather than automated scoring performance. Translation may have different effects on predictive scoring models than on neighborhood-based reliability auditing frameworks.

As multilingual representation models continue to improve, translation-free reliability auditing may provide a simpler, more scalable, and more inclusive approach for quality control in ILSAs.
\bibliography{custom}
\end{document}